\documentclass[conference,compsoc]{IEEEtran}
\usepackage{inputenc}           
\usepackage{graphicx}
\usepackage{epstopdf}
\usepackage{amsmath}            
\usepackage{amssymb}
\usepackage{amsfonts}
\usepackage{hyperref}           
\usepackage{subcaption}
\usepackage{caption}
\usepackage{xcolor}
\usepackage{multirow}
\usepackage{enumerate}
\usepackage[inkscapeformat=png]{svg}
\usepackage{makecell}
\usepackage{stfloats}

\newcommand{\RNum}[1]{\uppercase\expandafter{\romannumeral #1\relax}}

\ifCLASSOPTIONcompsoc
  \usepackage{cite}
\else
  \usepackage{cite}
\fi
\ifCLASSINFOpdf
\else
\fi
\usepackage{soul}

\newcommand{\ie}{\emph{i.e.,}}
\newcommand{\eg}{\emph{e.g.,}}
\newcommand{\cf}{\emph{c.f.,}}

\begin{document}
%
\title{MATE-Pred: Multimodal Attention-based TCR-Epitope interaction Predictor\textsuperscript{+}}



%
\author{
\IEEEauthorblockN{Etienne Goffinet\textsuperscript{*},
Raghvendra Mall,
Ankita Singh, 
Rahul Kaushik,
Filippo Castiglione}
\IEEEauthorblockA{Biotechnology Research Center\\
Technology Innovation Institute,
P.O. Box 9639, Abu Dhabi, U.A.E.\\ 
Emails: \{etienne.goffinet,raghvendra.mall,ankita.singh,rahul.kaushik,filippo.castiglione\}@tii.ae}}


\maketitle
\begingroup\renewcommand\thefootnote{*}
\footnotetext{Corresponding author}
\begingroup\renewcommand\thefootnote{+}
\footnotetext{Patent pending: U.S. Provisional Application No. 63/603,952}
\renewcommand{\thefootnote}{\arabic{footnote}}
\begin{abstract}
An accurate binding affinity prediction between T-cell receptors and epitopes contributes decisively to develop successful immunotherapy strategies.
Some state-of-the-art computational methods implement deep learning techniques by integrating evolutionary features to convert the amino acid residues of cell receptors and epitope sequences into numerical values, while some other methods employ pre-trained language models to summarize the embedding vectors at the amino acid residue level to obtain sequence-wise representations. 
Here, we propose a highly reliable novel method, MATE-Pred, that performs multi-modal attention-based prediction of T-cell receptors and epitopes binding affinity. The MATE-Pred is compared and benchmarked with other deep learning models that leverage multi-modal representations of T-cell receptors and epitopes. In the proposed method, the textual representation of proteins is embedded with a pre-trained bi-directional encoder model and combined with two additional modalities:  
a) a comprehensive set of selected physicochemical properties; 
b) predicted contact maps that estimate the 3D distances between amino acid residues in the sequences.
The MATE-Pred demonstrates the potential of multi-modal model in achieving state-of-the-art performance (+8.4\% MCC, +5.5\% AUC compared to baselines) and efficiently capturing contextual, physicochemical, and structural information from amino acid residues. The performance of MATE-Pred projects its potential application in various drug discovery regimes.
\end{abstract}

%
\IEEEpeerreviewmaketitle



\section{Introduction}
Lymphocytes T-cells (either CD4 T-helpers or CD8 T cytotoxic) are key actors of the \emph{adaptive} immunity  of vertebrates. 
T-cell receptors (TCR) are highly diverse cell membrane proteins that bind to fragments (called peptides) of the antigen that are presented by specialized antigen-presenting cells (APCs) such as macrophages and dendritic cells. 
In the initial phase of the immune response, the \emph{antigen} (\ie{} potentially pathogenic microorganism such as virus or bacteria) are captured by antigen-presenting cells (APCs) such as macrophages and dendritic cells, processed, and attached to the Major Histocompatibility Molecules (MHC). 
The peptide-MHC complex (pMHC) is exposed on the cell surface to be presented to the T-cells.
The interaction between the TCR and the pMHC that is required to activate the T-cells is highly specific and is a key step in the ignition of the immune response \cite{krogsgaard2005t,hudson2023can}. 
The binding affinity can be effectively determined by two short amino acid (AA) sequences \cite{krogsgaard2005t,davis1988t}: one is a peptide fragment of eight or more residues of the antigen that we here refer to as antigen-epitope or \emph{epitope}, and the second is its TCR counterpart \cite{nguyen2021pockets}. Within the TCR, the complementarity determining region 3 (CDR3) of TCR$\beta$ chain is known to be the imperative component that binds with its cognate epitope pairs \cite{krogsgaard2005t,davis1988t,xu2000diversity} as shown in Fig.\ref{fig:ref_tcr_epi}.

\begin{figure}[!ht]
    \centering
    \includegraphics[scale=0.30]{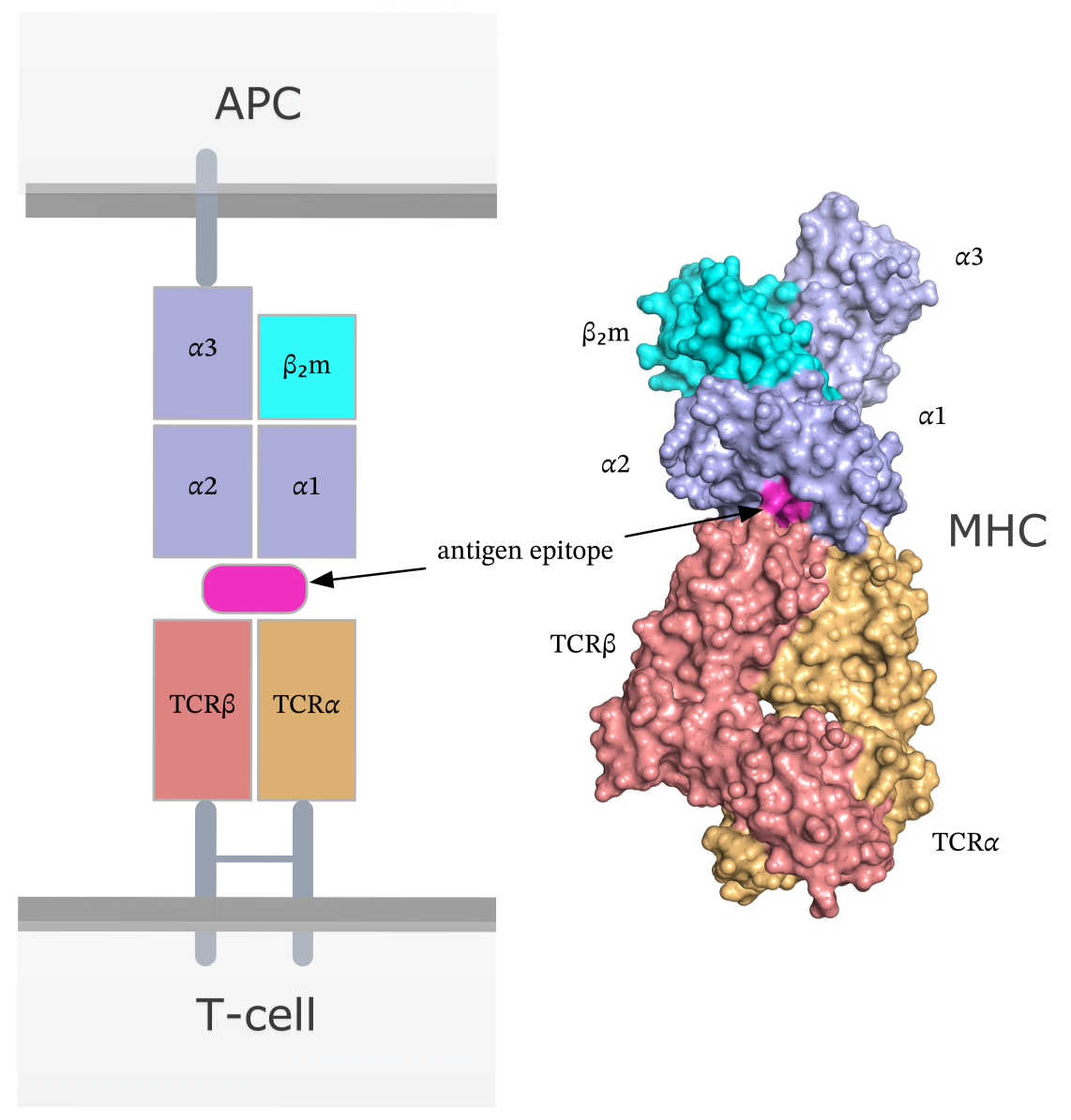}
    \caption{Crystal structure of the affinity-enhanced A3A TCR engaging with melanoma-
associated antigen 3 (MAGE-A3)-derived peptide presented by HLA-A*01 \cite{raman2016direct} (generated with data from \cite{raman2016direct} and visualized with PyMOL).}
    \label{fig:ref_tcr_epi}
\end{figure}

As proteins, both T-cell receptors (TCRs) and epitopes possess unique structural organization, leading to the need for other representations or ``modalities'' in their representation.
The primary structure of proteins refers to the linear arrangement of amino acid residues in the polypeptide chains. 
The secondary structure involves the folding of the polypeptide chain into regular structures such as alpha-helices and beta-strands, that are stabilized by hydrogen bonds between amino acid residues \cite{sun2004overview}. The tertiary structure encompasses the three-dimensional conformation
resulting from interactions between different regions and the polypeptide chain's second structural elements (helices and strands). For TCRs, the tertiary structure includes the arrangement of their variable and constant domains which is a key determinant in antigen recognition. Epitopes, on the other hand, fold into the specific conformations dictated by the interactions between their amino acid residues. The tertiary structures of TCRs and epitopes are vital for their proper functioning and interactions during immune responses, where TCRs recognize the antigen peptides presented by APCs.

Accurate estimation of TCR-epitope binding affinity is critical to unraveling the underlying biological mechanisms that are imperative for surveillance and response to disease, and developing novel vaccines or immunotherapies. 
Developing an accurate \textit{in silico} framework is indispensable in automating the screening process for identifying suitable T-cell receptors (TCRs) that recognize a specific epitope of interest. Such frameworks offer a potential alternative to the traditional wet lab assays by minimizing the time, costs, and experimental limitations \cite{hudson2023can}. By enabling rapid and efficient screening, this approach facilitates expedited development of personalized T-cell therapies \cite{graham2018cancer,zhao2019engineered}.

In recent years, numerous Machine Learning (ML) techniques have been utilized to estimate the binding affinity between TCR and epitope sequences \cite{gielis2019detection, jokinen2021predicting, jurtz2018nettcr, montemurro2021nettcr, springer2020prediction, weber2021titan, cai2022atm,wu2021tcr}. For instance, models such as TCRex \cite{gielis2019detection} and TCRGP \cite{jokinen2021predicting} make use of Random Forests and Gaussian processes, respectively, while
more recent models focus on utilizing the capabilities of Deep Neural Networks.
Models such as NetTCR \cite{jurtz2018nettcr} and NetTCR2.0 \cite{montemurro2021nettcr} are built using convolutional neural networks (CNN \cite{lecun1995convolutional}) with varying kernel sizes to encode both TCR and epitope amino acid sequences followed by multiple linear layers to estimate the binding affinity. In a different context \cite{moris2021current}, ImRex uses the CNN architecture to predict TCR-epitope binding based on the interaction map representation extracted from physicochemical properties (\eg{} hydrophobicity, isoelectric point, mass, and hydrophilicity).

To capture the contextual information in the AA sequences, ERGO \cite{springer2020prediction} 
used a long-short term memory (LSTM \cite{hochreiter1997long}) neural network. Similarly,
TITAN \cite{weber2021titan} and ATM-TCR \cite{cai2022atm} focused on attention mechanisms in
their deep neural network architectures. 
A majority of these deep learning models \cite{jokinen2021predicting, gielis2019detection, jurtz2018nettcr, montemurro2021nettcr} maps each AA in the ligand (\ie{} TCR or epitope) sequence to a real-valued vector representation.
using evolution-based BLOSUM matrices \cite{henikoff1992amino}. The BLOSUM (BLOcks SUbstitution Matrix) matrix is a \emph{substitution} is a set of amino acid substitution log of odds scores derived from the ungapped multiple sequence alignment of proteins \cite{henikoff1992amino}. BLOSUM matrices are based on local alignments and are employed to assign scores to alignments between protein sequences that have undergone evolutionary divergence. 
However, models using BLOSUM-based embeddings face challenges in terms of limited performance, especially when predicting the binding affinity for previously unseen (novel) epitopes that are not part of the training dataset \cite{cai2022atm}.

As an alternative approach, various amino acid embedding models have been introduced \cite{springer2020prediction, wu2021tcr, sidhom2021deeptcr}. These models leverage large datasets of unpaired T-cell receptor (TCR) sequences and employ an autoencoder framework, where the input sequence is used for self-supervision to train and learn the embeddings.
Among these models, BERT-TCR \cite{wu2021tcr} and DeepTCR \cite{sidhom2021deeptcr} were inspired by language models such as BERT \cite{devlin2018bert} and ELMo \cite{peters2018deep} respectively, and have shown to learn more efficient contextualized embeddings for TCR and epitope sequences with improved generalization performance. 
The major limitation of these models include the generation of larger-sized embedding vectors than BLOSUM-based methods. In an attempt to address this issue, average pooling has been commonly employed as an approach to reduce the computational burden and size of these embeddings. However, it has been empirically demonstrated, as indicated in PiTE \cite{zhang2022pite}, that this pooling approach can result in the degradation of position-specific information encoded in the language models, decreasing their predictive performance.

In contrast to this significant amount of \emph{uni-modal} binding prediction models (\ie{} AA string, BLOSUM embedding, or physicochemical features), there have been, to the best of our knowledge, no attempts to leverage their combinations. 

It is worth mentioning that the proposed method MATE-Pred renders a very different strategy compared to the previously purposed multi-modal approaches \cite{luu2021predicting, moris2021current}, where the TCR and epitope are seen as two modalities of the TCR-pMHC complex. Instead, MATE-Pred considers the TCR and epitope as two AA sequences and several modalities describe each of them.
The novelty of the present work is, therefore, to evaluate the effectiveness of multi-modal representations of TCR and epitope for predicting their binding affinity.
%
In particular, in this paper, we
\begin{enumerate}[I]
    \item Introduce MATE-Pred, a new multi-modal architecture for TCR-epitope binding affinity prediction
    \item Demonstrate the significant impact of multi-modal combinations, and benchmark different multi-modal settings on several datasets
    \item Illustrate that MATE-Pred model outperforms the state-of-the-art (SOTA) on an independent test set comprising epitopes and TCRs from less studied species.
\end{enumerate}
Finally, we provide the model training source code and weights, as well as pre-processed datasets, as a baseline for future research\footnote{https://github.com/anom-243234d/MATE}.

\section{Materials and Methods}\label{sec:method}
\subsection{Training Dataset Collection}
We used the database of positive and negative examples of TCR-epitope pairs as collected by the PiTE model \cite{zhang2022pite}. 
In details, we obtained TCR-epitope pairs with binding affinity scores from three publicly available databases, VDJdb \cite{shugay2018vdjdb}, IEDB \cite{vita2019immune}, and McPAS \cite{tickotsky2017mcpas}. 
Only pairs with MHC Class I  epitopes and TCR$\beta$ CDR3 sequences were included in the training set. Sequences retrieved from these three databases were screened for presence of any non-standard amino acid residue and duplicates to compile a total of 150,008 non-redundant TCR-epitope pairs that are known to bind.

To achieve a balanced dataset with an equal number of positive and negative samples, we utilized TCR sequences from the TCR repertoires of healthy controls sourced from the ImmunoSEQ portal \cite{nolan2020large}. The TCRs of the known binding TCR-epitope pairs were randomly substituted with TCRs from the healthy control dataset. Combining these positive pairs with the generated negative pairs, our dataset consisted of 300,016 unique TCR-epitope pairs.

\subsection{Independent Test Set} \label{sect:indep_test_set}
We downloaded 80,936 TCR-epitope pairs from VDJdb (accessed on May 10, 2023).It may be noted that the VDJdb is constantly growing in number of available TCR-epitope pairs.
To create an independent test set, we selected the unique epitopes present in VDJdb that did not overlap with our training dataset. This led to 842 distinct epitopes (744 MHC class I and 98 MHC class II) spanning over 24 different pathogens, including SARS-CoV-2, different strains of influenza, HIV, and rare epitopes from E. Coli, Pseudomonas, Epstein-Barr virus, etc. resulting in a total of 10,171 positive TCR-epitope binding pairs (8,007 TCR-epitope pairs for MHC Class I epitopes and 2,164 TCR-epitope pairs for MHC Class II epitopes). 
To create a balanced test set, we generated negative samples using the 842 epitopes in combination with TCR sequences from healthy controls in the ImmunoSEQ portal leading to the final test set of 20 342 TCR-epitope pairs. 
We regarded this dataset as challenging and intriguing as it includes rare epitopes and MHC Class II epitopes, that imparts an additional level of complexity to the task.

\subsection{TCR and Epitope modalities}
In the proposed approach, both the T-cell receptor and epitopes are treated as amino acid sequences. The same pre-processing and multi-modal extraction process was applied to both. Therefore, the following description of data pre-processing refers to AA sequences, encompassing both TCRs and epitopes interchangeably, referred to as \emph{ligand}.

\subsubsection{AA sequence}
In line with recent literature \cite{zhang2022pite}, the first modality we consider is the AA primary sequence as a string, embedded with a pre-trained encoder model. Following the recommendations of \cite{zhang2022pite}, the same ELMo base representation \cite{zhang2023context} is used to develop the proposed method. This embedding model is based on a bi-directional Long Short-Term Memory network (LSTM) that leverages the amino acid context in the whole string. As the attention-based model requires equal-length AA sequences, every TCR and epitope sequence is truncated or padded to a context length of 22, which is an upper bound of our datasets' sequences length. The embedding step transformed each AA into a real-values vector of size 1024. Thus, each ligand sequence is converted into a size $22 \times 1024$ matrix.

\subsubsection{Physicochemical Features}
As the second modality, the AA sequences are described with a representation of the physicochemical features through a set of selected descriptors. We were deliberately liberal in our physicochemical feature selection in an attempt to capture most of the available information and considering that our target architecture can perform feature recombinations (\cf{} Sub-Sec.~\ref{ssec:multimodalfusion}). These features include physicochemical descriptors as well as global properties and were extracted using the \emph{peptides} package (v0.3.1) in Python (v3.7.12) \cite{peptides2022}.

The physicochemical descriptors ($n_{c}=12$) include the following features (represented in a 75-dimensional space):
BLOSUM indices, 
Cruciani properties, 
FASGAI vectors, 
Kidera factors, 
MS-WHIM scores, 
PCP descriptors, 
ProtFP descriptors, 
Sneath vectors, 
ST-scales, 
T-scales, 
VHSE-scales and 
Z-scales, detailed in Appendix Table \ref{table:physicochemical_list}.

These features further include global properties ($n_{p}=13$) such as 
aliphatic index, 
autocorrelation, 
autocovariance, 
Boman instability index, 
charge, 
hydrophobic moment $\alpha$, 
hydrophobic moment $\beta$, 
hydrophobicity, 
instability index,
isoelectric point, 
mass shift, 
molecular weight and 
mass over charge ratio, as detailed in Appendix Table \ref{table:properties_list}. In total, the obtained set of physicochemical and global properties has a dimension of 88.
%
\vspace{-3mm}
\subsubsection{Contact Map}
For the last modality, the predicted contact map (C-map) representation \cite{wang2017accurate} was accounted as a surrogate for 
the tertiary / 3D structure of a TCR/epitope complex. The main rationale for selecting the contact maps  was two-fold: it was both lightweight (which helps control the computation cost) and invariant to the rotation or translation. 
For a given AA sequence of a TCR/epitope, this map is a matrix that contains, in each cell $(i,j)$, an estimate of the distance between two residues ($i$ and $j$). An illustration is displayed in Fig.\ref{fig:contact_map}. We used the unsupervised self-attention contact map predictions from the pre-trained language model (ESM-2), version with 650M parameters \cite{lin2023evolutionary}, available at \url{https://github.com/facebookresearch/esm}. We padded the C-maps of small sized epitopes with 0's, \ie{} no-contacts. 
Finally, given that the C-maps are distance matrices and symmetric, the upper triangular is enough to keep all information, which for an AA sequence of size $\ell=22$, resulted in a $\ell(\ell+1)/2 = 253$-long numeric vector. 

\begin{figure}[!ht]
    \centering
    \includegraphics[scale=0.55]{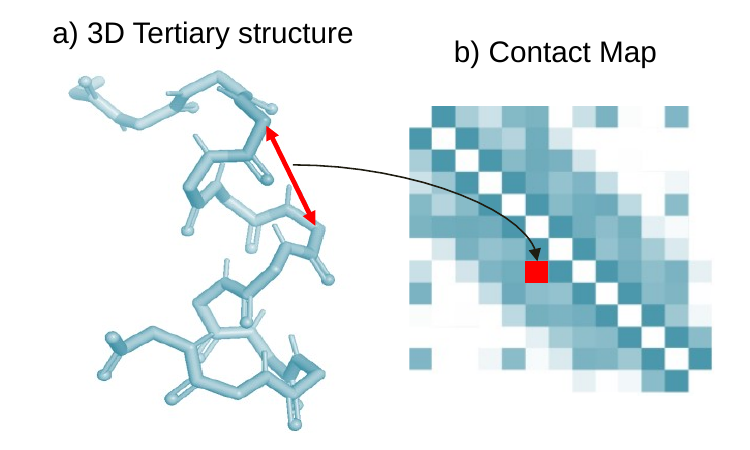}
    \caption{a) Tertiary / 3D structure of an AA sequence; b) The contact map is a distance $\ell\times \ell$ matrix where $c_{ij}$ value is the distance between the amino acid in position $i$ and $j$ in the 3D space representation.}
    \label{fig:contact_map}
\end{figure}

\subsection{Multi-Modal architecture}
\subsubsection{Modality Fusion}\label{ssec:multimodalfusion}
Inspired by the recent literature on the topic \cite{xu2023multimodal}, we bench-marked different multi-modality fusion methods (\cf{} Sec.~Appendix). We consequently select to focus on the early concatenation scheme (also known as "all-attention"), which presents the best consensus between computation cost and prediction performances. 
This method consists of concatenating the AA sequence embeddings to the other modalities' projections before the encoder blocks. It has the advantage of integrating the whole multi-modal context in one step. Given the quadratic complexity of the encoder block, this strategy might seem computationally expensive, but this cost increase remains marginal as we keep the modality projection dimension low. 

In this setting, each modality is fed to two Linear layers, interleaved with a non-linear ReLU activation function, and projected onto a 1024-d vector. This first step aims at re-combining the input features, which we expected to be redundant (\eg{} because of our liberal physicochemical feature selection or the spatial dependency between AA on the contact maps).
This vector is then concatenated to the AA sequence embeddings to obtain a 24x1024 tensor representation. This fusion's process is depicted in panel \emph{a} of Fig.\ref{fig:multimodalencoder}, and is followed by the Attention Block.

\subsubsection{Attention-based encoder}
The attention-based encoder block is a key component of our proposed architecture. Now standard in natural language processing literature, this model corresponds to the encoder part of the Transformer architecture \cite{vaswani2017attention}, and combines a self-attention mechanism with a feed-forward network to obtain a context-aware representation of its input. 
Formally, given an input tensor $X$ with dimension $\ell\times d$, the self-attention is defined by:
\begin{equation}
Attn(X) = {\rm SoftMax}(Q.K^T / \sqrt{d_k}) V,
\end{equation}
where the matrices $Q$, $K$, and $V$ share the same dimensions $1024\times 1024$ and are linear projections of the same input matrix $X$. 
Intuitively, this process estimates the compatibility between the queries Q and the keys K and uses this compatibility score to weight the values V. 
Following the original transformer architecture \cite{vaswani2017attention}, we use multi-head attention, where this mechanism is repeated in parallel several times (in our case, two heads). This attention part is then followed by a Feed-Forward network that finally outputs the attention-aware representation of the AA sequences. This process has a quadratic complexity w.r.t the context length. 

Combined, the modality fusion block and the attention-based encoder block form the AA sequence multi-modal encoder, depicted in panel \emph{a} of Fig.\ref{fig:multimodalencoder}. Our proposed architecture contains two separate multi-modal AA sequence encoders: the TCR encoder and the Epitope encoder, which are depicted in the panel \emph{b} of Fig.\ref{fig:multimodalencoder}.

\begin{figure*}[!ht]
    \centering
    \includegraphics[scale=0.6]{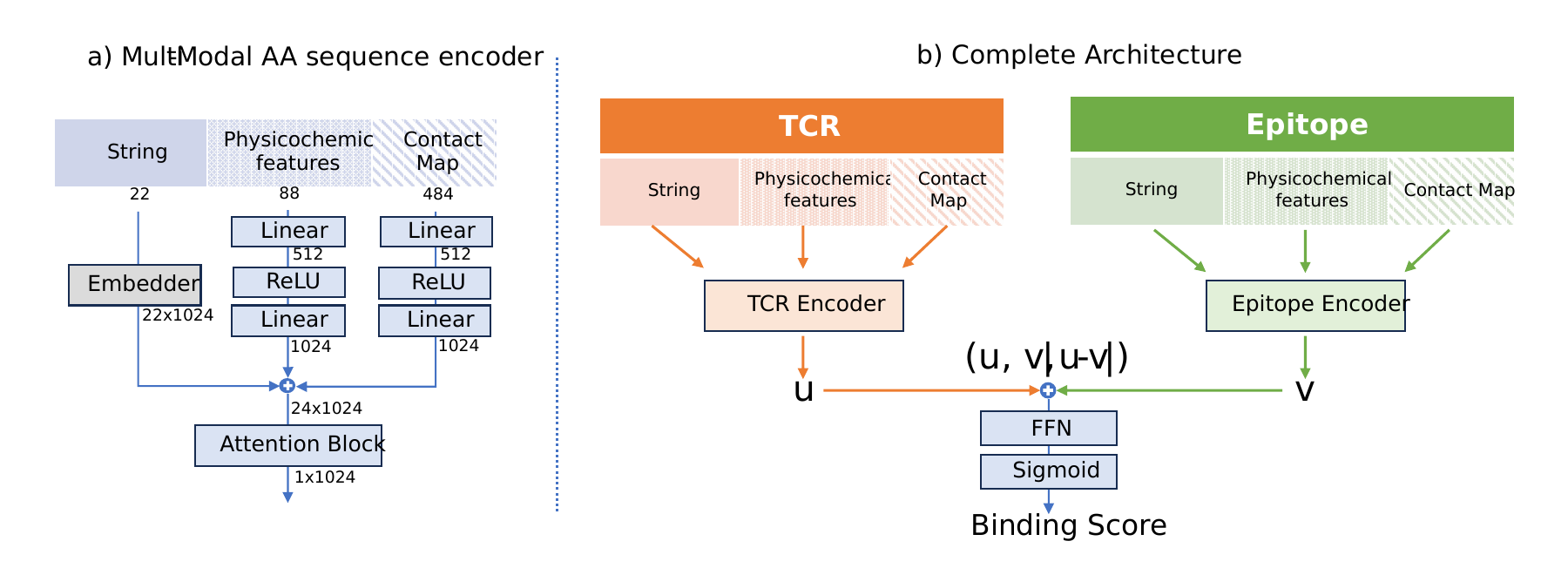}
    \caption{a) Multi-modal AA sequence encoder. The grey \emph{embedder} block on the left is pre-trained and its weights are fixed. b) the full architecture, featuring two Multi-modal AA sequence encoders (\emph{TCR Encoder} and \emph{Epitope Encoder}) and the final Feed-Forward Network block.}
    \label{fig:multimodalencoder}
\end{figure*}

\begin{figure*}[t]
    \centering
    \includegraphics[scale=0.7]{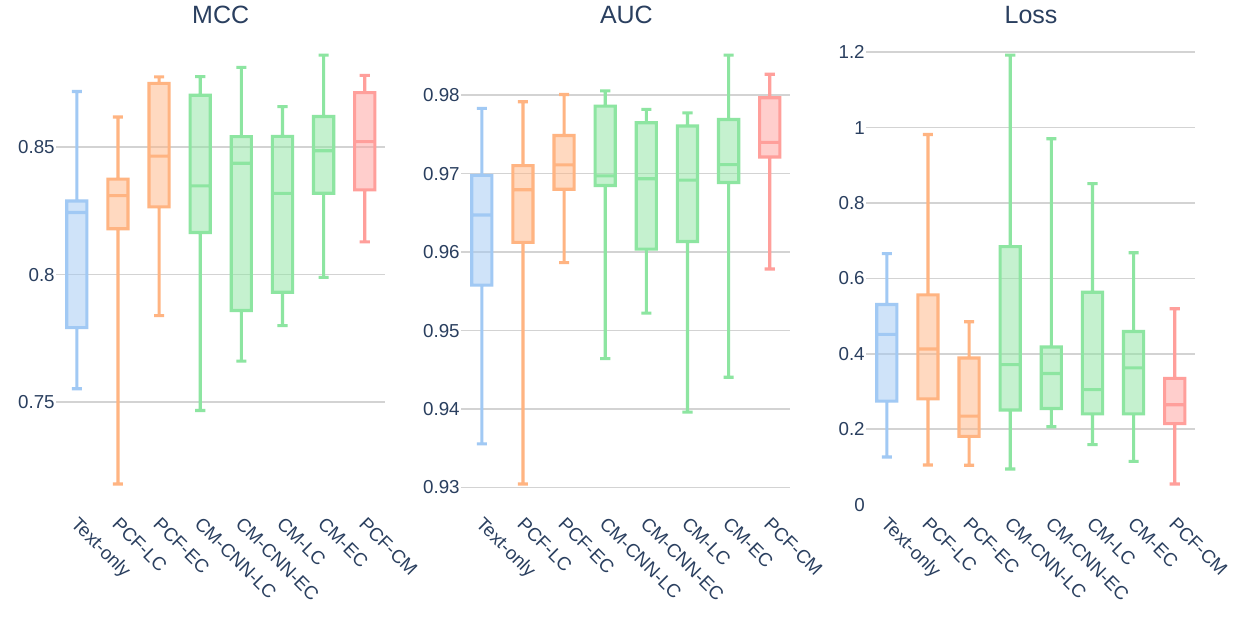}
    \caption{Multi-Modal benchmark results. From left to right: \emph{Text-only}: uni-modal architecture; \emph{PCF-LC}: Text + Physicochemical feature (late concat); \emph{PCF-EC}:  Text + Physicochemical features (early concat); \emph{CM-CNN-LC}:  Text + Contact Map (CNN and late concat); \emph{CM-CNN-EC}:  Text + Contact Map (CNN and early concat); \emph{CM-LC}: Text + Contact Map (late concat); \emph{CM-EC}: Text + Contact Map (early concat); \emph{PCF-CM}: Text + Physicochemical features (early concat) + Contact Map (early concat).}
    \label{fig:benchmark_results}
\end{figure*}

\subsubsection{Final projection}
Following \cite{zhang2022pite}, the outputs of these encoders (respectively denoted $u$ and $v$) are then concatenated together and with their absolute difference $|u-v|$. The addition of this last element is meant to help the model focus on the difference between the TCR and the epitope representations. After concatenation, the output dimension is 3072.
The last block is a Feed-Forward neural network composed of a linear layer $(3072 \rightarrow 1024)$, a Batch Normalization Layer, A dropout (0.3), and finally a linear layer $(1024 \rightarrow 1)$ that outputs the binding affinity score.  

The complete architecture features 22M trainable parameters, with parameter details highlighted in Table \ref{table:param_count}.

\begin{table}[]
\caption{Architecture dimensions details.}
\begin{tabular}{llll}
\hline
\multicolumn{1}{|l|}{Block}        & \multicolumn{1}{l|}{Module}             & \multicolumn{1}{l|}{Sub-Module} & \multicolumn{1}{l|}{Size} \\ 
\hline \hline \hline
\multicolumn{1}{|l|}{\multirow{4}{*}{\thead{Multi Modal \\ TCR Encoder}}}     & \multicolumn{2}{l}{Physicochemical (Phi) Projection}     & \multicolumn{1}{l|}{570880}        \\ \cline{2-4} 
\multicolumn{1}{|l|}{}             & \multicolumn{2}{l}{C-map Projection}    & \multicolumn{1}{l|}{655360}        \\ 
\cline{2-4} 
\multicolumn{1}{|l|}{}             & \multicolumn{1}{l|}{\multirow{2}{*}{Attention Encoder}} & Self-Attention     & \multicolumn{1}{l|}{8397824}       \\ 
\cline{3-4}
\multicolumn{1}{|l|}{}             & \multicolumn{1}{l|}{}       & FFN        & \multicolumn{1}{l|}{68640}         \\ \hline \hline
\multicolumn{1}{|l|}{\multirow{4}{*}{\thead{Multi Modal \\ Epitope Encoder}}} & \multicolumn{2}{l}{Physicochemical (Phi) Projection}  & \multicolumn{1}{l|}{570880}  \\ \cline{2-4} 
\multicolumn{1}{|l|}{}            & \multicolumn{2}{l}{C-map Projection}     & \multicolumn{1}{l|}{655360}        \\ \cline{2-4} 
\multicolumn{1}{|l|}{}            & \multicolumn{1}{l|}{\multirow{2}{*}{Attention Encoder}} & Self-Attention    & \multicolumn{1}{l|}{8397824}       \\ \cline{3-4} 
\multicolumn{1}{|l|}{}            & \multicolumn{1}{l|}{}    & FFN   & \multicolumn{1}{l|}{68640}         \\ 
\hline \hline
\multicolumn{1}{|l}{Final Projection}      
                                & \multicolumn{2}{l}{}   & \multicolumn{1}{l|}{3149825}       \\ \hline
                                &                        & Total:          & 22535233           \\ \hline       

\end{tabular}
\label{table:param_count}
\end{table}

\section{Experimental setup}\label{sec:exp_setup}

\subsection{Implementation}

We tested different possible multi-modality combinations: 
1) \emph{Text-only}: the unimodal architecture based on AA string representation (comparable to PiTE architecture); 
2) \emph{PCF}: text and physicochemical features; 
3) \emph{CM}: text and contact map; 
4) \emph{PCF-CM}: all modalities: Text, physicochemical features and contact map. 
In addition, we also show the results of different multi-modality fusion methods: the early concatenation (where the modalities are aggregated before the attention block, denoted with the suffix \emph{-EC}) and the late concatenation (where the modalities are aggregated after the attention block, denoted with the suffix \emph{-LC}). For the CMAP, in addition to the final architecture depicted in Fig.\ref{fig:multimodalencoder}, we considered another variant that handles the contact map as an image and substitutes the initial linear projection layers with 2D convolution layers.

The training is implemented in Pytorch and optimizes the binary cross-entropy loss with Adam $(lr =5\text{e-}3, \beta_1 = 0.9, \beta_2 = 0.98, \epsilon =1\text{e-}9)$, a batch size of 512, and considered fully completed after 200 epochs. In this setting, less than 4 hours are required to train the multi-modal model on an NVidia RTX A6000 GPU. Our training runs were performed on a heterogeneous cluster of instances featuring Nvidia RTX A6000, RTX 3070, and Tesla T4 GPUs.

\subsection{Evaluation}
In addition to the binary cross-entropy loss and area under the curve (AUC), we also considered the Matthews correlation coefficient (MCC) score, 
which has been shown to be more reliable than accuracy or F1 score \cite{chicco2020advantages} as it uses all components of the binary confusion matrix. 
Formally, this score is defined as:
\begin{align*}
    & MCC = \\
    & \frac{ TP \times TN - FP \times FN} {\sqrt{ (TP + FP) ( TP + FN) ( TN + FP ) ( TN + FN) } }
\end{align*}
Given the significant epitope redundancy in the dataset (there are 982 unique epitopes in our 300k samples) and following \cite{zhang2022pite}, we based our dataset train/test/validation split on the epitope distribution, such that epitopes present in one of the splits was not present in the others. 
However, contrary to \cite{zhang2022pite}, we used 10 different splits to evaluate the dataset instead of one, which is a more reliable cross-validation process as it allows a split-independent evaluation. For each score, we report the test value observed at the epoch that maximizes the corresponding validation score.

\subsection{Independent dataset validation}
We used the independent test set (\cf{} Sec.~\ref{sect:indep_test_set}) to compare our model to existing SOTA baselines. We emphasize that this dataset contains MHC Class I and Class II epitopes and that there is no overlap between this 20k-sample test set and our 300k-sample training set, making this test particularly challenging for most methods.

We report the results obtained with our method trained and validated on the original 300k dataset (with ten different train/valid splits) and tested on the 20k independent test. In this setting, the samples from the independent test dataset are never seen during training.

We considered four baseline models for benchmarking: ATM-TCR \cite{cai2022atm}, TEINet \cite{jiang2023teinet}, NetTCR \cite{jurtz2018nettcr} and ImRex \cite{moris2021current}. 

ATM-TCR\footnote{Based on https://github.com/Lee-CBG/ATM-TCR} is a SOTA model with an architecture close to ours, as it also contains separate attention-based encoders for the TCR and Epitope, but doesn't support multi-modality. This model was trained on the same dataset as Pite and our proposed model, thus, making it a pertinent baseline to assess the advantage of multi-modal representations. 

The second baseline that we considered is TEINet\footnote{Based on https://github.com/jiangdada1221/TEINet}, trained on a custom extraction of 44k samples from VDJdb, McPAS, \cite{lu2021deep}, and based on an autoregressive model with gated recurring units \cite{jiang2023deep} pre-trained on a large extraction from ImmunoSeq \cite{emerson2017immunosequencing}.

The third baseline that we considered is NetTCR \footnote{Based on https://github.com/mnielLab/netTCR}, which is based on a shallow CNN architecture and has been trained on a dataset of 70k samples from IEDB. We used
the NetTCR version 1.0 as our dataset only contains CDR$\beta$, whereas NetTCR's last version requires both CDR$\alpha$ and $\beta$.

Finally, the last baseline that we considered is ImRex \footnote{Based on https://github.com/pmoris/ImRex} which predicts the TCR-Epitope binding by estimating the interaction map using the physicochemical features, and using 2D CNN on these maps. This model was trained on the August 2018 release of VDJdb. 

\section{Results and Discussion}\label{sec:results}

\subsection{Benchmark}

The results of the Multi-Modality benchmark are displayed in Fig.\ref{fig:benchmark_results}. For all the metrics, the multi-modal approach shows a significant improvement over the uni-modal text baseline (comparing \emph{Text-only} and \emph{PFC-CM}) on this dataset.
The MCC score confirms the prevalence of the early concatenation
strategy for both modalities. With either concatenation strategy, the CNN handling of the contact map (methods \emph{CM-CNN-LC}, \emph{CM-CNN-EC}) is underperforming. This can possibly be attributed to the sparse contacts in TCR and epitope sequences.
Moreover, each modality addition improves predictive performance. The performance gain is roughly the same for each modality (comparing \emph{PCF-EC} and \emph{CM-EC}), with a minor MCC and AUC advantage for the Contact Map variant.
However, their combination produces the best model, suggesting that each modality improves the performance on a distinct part of the dataset. 
To test the impact of pre-trained embeddings, we also experimented with a variant of the model (not shown on Fig.\ref{fig:benchmark_results}) where the embedding layer is trained from scratch and in combination with the standard sinusoidal positional encoding. This variant performs very poorly, with an average test MCC of 0.28 and test AUC of 0.69, confirming the importance of the pre-trained embedding for these tasks, as it was expected \cite{zhang2022pite}.

\subsection{Independent Test Data}
\vspace{-2mm}
The results obtained on the challenging independent test set are displayed in Fig.\ref{fig:indep_test_results}. The relatively modest scores (overall less than 0.35 MCC and 0.75 AUC) are not surprising, considering the difficulty of our target dataset. Among the baseline methods, only TEINet seemed to grasp a signal (with an MCC of 0.164 and AUC of 0.585), while the other three unimodal methods performed only slightly above a random guess. We managed to reproduce and explain these low scores, by considering the unimodal text-only version of our model and removing the pre-trained embedding part. This variant (called "Text-only WPE" in Fig.\ref{fig:indep_test_results}) reproduces an architecture close to ATM-TCR and produces the same performance level. This confirms the impact of pre-trained embedding \cite{zhang2023context,zhang2022pite}, which has already been established in the TCR-epitope binding prediction context and was observed in our 300k-sample dataset benchmark. This also explains TEINet's good performance, as this model uses pre-trained embedding.

The multi-modal architectures exhibit a different behavior on this dataset than in the previous benchmark. Compared to the text-only model (MCC 0.206, AUC 0.624), the addition of physicochemical characteristics (\emph{PCF-EC}) shows only a moderate (although positive) effect on the predictions (MCC 0.218, AUC 0.626). However, the contact maps addition (\emph{CM-EC}) has a significant impact, achieving an average of 0.289 MCC and 0.676 scores, representing +0.083 MCC and +0.052 AUC compared to the text-only model. 

Combining both modalities increases the average performance again, although by a small margin, achieving 0.29 MCC and 0.679 AUC scores, representing +0.084 MCC and +0.055 AUC points increase compared to the text-only model. This modest increase makes sense given the low performance of \emph{PCF-EC}. However, the improvement becomes much more evident when considering the best-obtained scores rather than the average. With maximum values of 0.355 MCC and 0.742 AUC, this represents an increase of +0.135 MCC and +0.124 AUC points. 

\begin{figure*}
    \centering
    \includegraphics[scale=0.7]{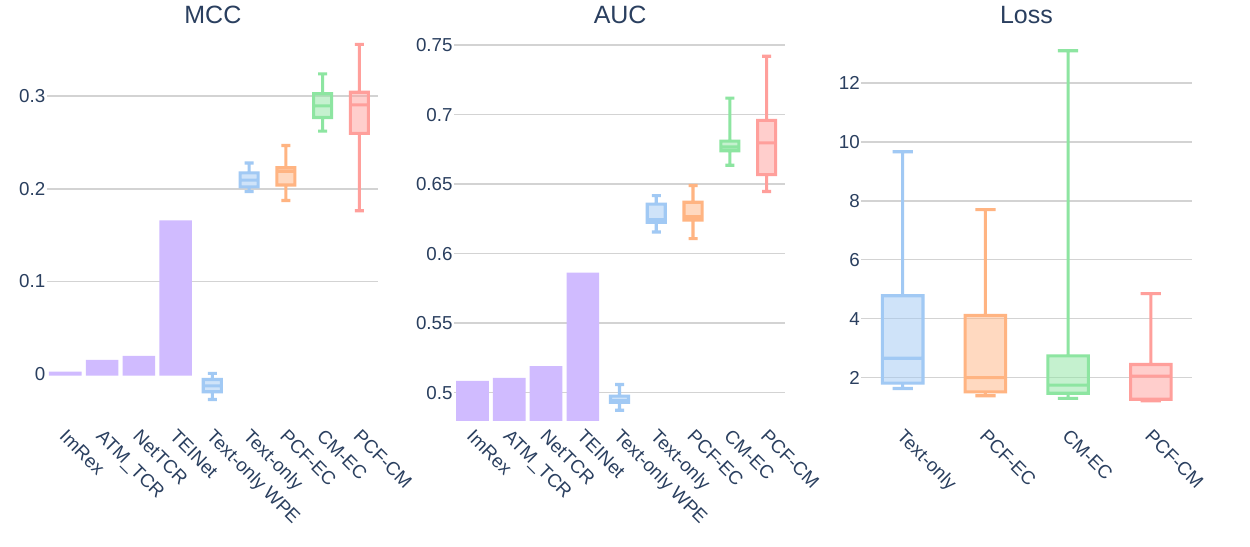}
    \caption{Results on the independent test dataset; Text-only WPE is a variant of Text-only with AA string sequences embedding trained from scratch.}
    \label{fig:indep_test_results}
\end{figure*}

Overall, these results confirm the importance of pre-trained embedding for TCR-Epitope embedding and show the impact of multi-modal representation on generalization performance on out-of-the-box samples. We display on Fig.\ref{fig:confusion_matrices} the details of the confusion matrices, by MHC Class, of the \emph{Text-only} and \emph{PCF-CM} models with the highest validation MCC score. These detailed view show that the multi-modality improves every part of the MHC Class I confusion matrix. In contrast, the benefit on the MHC Class II subset seems to mainly impact the negative binding prediction. 

\begin{figure}[t!]
    \centering
    \includegraphics[scale=0.4]{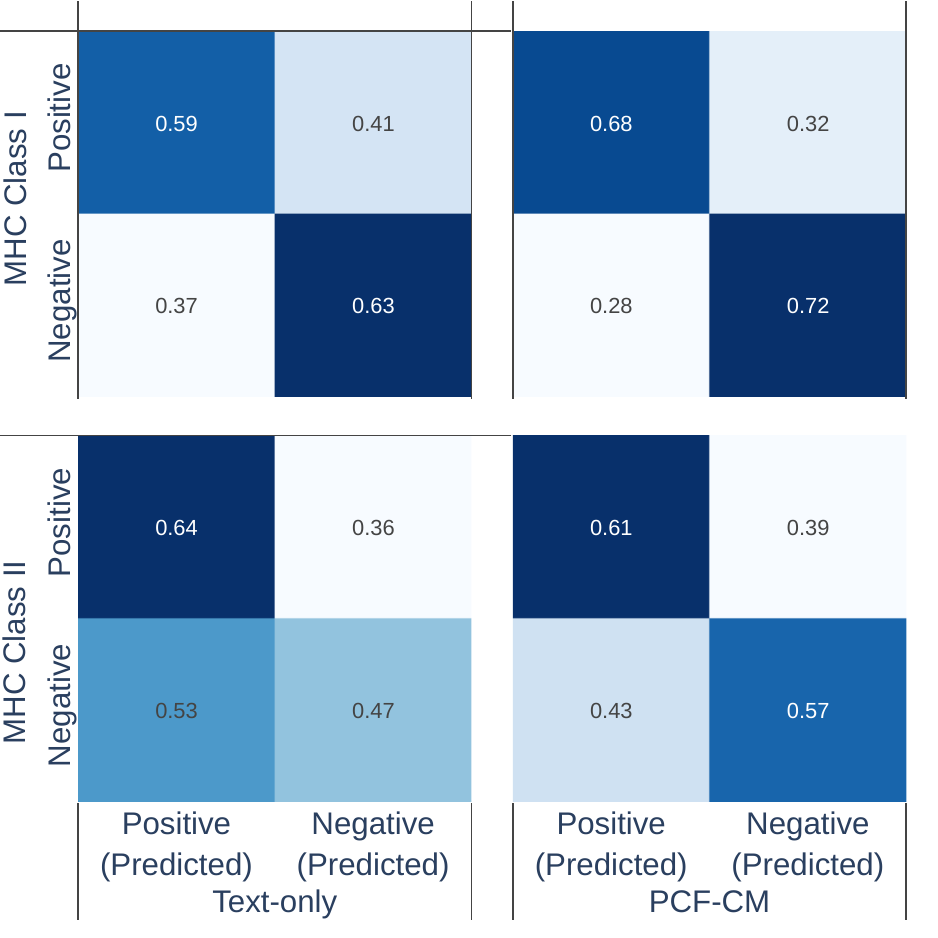}
    \caption{Confusion matrices derived by MHC class and method (\emph{Text-only}, \emph{PCF-CM}).}
    \label{fig:confusion_matrices}
\end{figure}

\section{Discussion \& Conclusion}
Our study tackles the challenge of predicting TCR-epitope binding affinity by introducing and comparing various attention-based architectures that use multi-modal representations.
While previous methods have predominantly focused on individual modalities, MATE-Pred incorporates textual data, selected physicochemical properties, and predicted contact maps to capture a broader range of information. Through a comprehensive ablation study, we have demonstrated the effectiveness of the proposed multi-modal prediction model. MATE-Pred demonstrates superior performance over the transformer-like sequence-encoder model and establishes a new state-of-the-art performance on an independent test set. 

Our findings underline the importance of multi-modal integration in TCR-epitope binding affinity prediction, as it significantly improves the predictive performance. By incorporating diverse information sources, our approach lays the foundation for future advancements in predicting the immunogenicity of antigenic peptides. In this regard, it opens opportunities to explore personalized (\ie{} patient-specific) immunotherapy strategies and develop tailored treatment options based on more accurate predictions.

Moving forward, we anticipate further advancements in multi-modal deep learning models for TCR-epitope interactions. By incorporating features from tertiary structures as well as the integration of additional modalities, such as gene or protein expression data, could provide even more comprehensive insights.
The combination of our approach with experimental validation holds promise for advancing immunotherapy and ultimately improving patient outcomes.

\small{
\bibliographystyle{IEEEtran}
\bibliography{IEEEabrv,main}
}
\newpage
\appendix

\setcounter{table}{0}
\renewcommand{\thetable}{A\arabic{table}}
\vspace{-4mm}
\begin{table}[!ht]
\caption{Ligand Physicochemical descriptors details.}
\centering
\begin{tabular}{|c|c|}
\hline
\textbf{Name} & \textbf{Dimensions} \\ \hline
BLOSUM Indices & 10 \\ \hline
Cruciani Properties & 3 \\ \hline
FASGAI vectors & 6 \\ \hline
Kidera factors & 10 \\ \hline
MS-WHIM scores & 3 \\ \hline
PCP properties & 5 \\ \hline
ProtFP descriptors & 8 \\ \hline
Sneath vectors & 4 \\ \hline
ST-scales & 8 \\ \hline
T-scales & 5 \\ \hline
VHSE-scales & 8 \\ \hline
Z-scales & 5 \\ \hline
\hline
Total length  & 75 \\  \hline                                               
\end{tabular}
\label{table:physicochemical_list}
\end{table}

\begin{table}[!ht]
\caption{Ligand Global Properties details.}
\centering
\begin{tabular}{|c|c|}
\hline
\textbf{Name} & \textbf{Dimensions} \\ \hline
Aliphatic Index & 1 \\ \hline
Autocorrelation & 1 \\ \hline
Autocovariance & 1 \\ \hline
Boman Index & 1 \\ \hline
Lehninger Charge & 1 \\ \hline
Hydrophobic Moment $\alpha$ & 1 \\ \hline
Hydrophobic Moment $\beta$ & 1 \\ \hline
Hydrophobicity & 1 \\ \hline
Instability Index & 1 \\ \hline
Isoelectric Point & 1 \\ \hline
Mass Shift & 1 \\ \hline
Molecular Weight & 1 \\ \hline
Mass over charge ratio & 1 \\ \hline
\hline
Total length  & 13 \\  \hline                                               
\end{tabular}
\label{table:properties_list}
\end{table}

\end{document}